\documentclass[runningheads]{llncs}
\usepackage{graphicx}
\usepackage{cite}
\usepackage{tabularx}
\usepackage{enumerate}
\usepackage{bigstrut}
\usepackage{dsfont}
\usepackage{float}
\usepackage[hidelinks,colorlinks,allcolors=black,pdfstartview=Fit]{hyperref}
\usepackage{amsmath,amssymb}
\usepackage{mathrsfs}
\usepackage[ruled,vlined]{algorithm2e}
\usepackage{booktabs}
\usepackage{algpseudocode}
\usepackage{multirow}
\usepackage{ulem}
\usepackage{changes}
\usepackage{comment}
\usepackage{subcaption}
\usepackage{marvosym}

\begin{document}
\title{CANAMRF: An Attention-based Model for Multimodal Depression Detection}
\titlerunning{An Attention-based Model for Multimodal Depression Detection}
%
\author{Yuntao Wei \and Yuzhe Zhang \and Shuyang Zhang \and Hong Zhang\textsuperscript{\Letter}}
\authorrunning{Y. Wei et al.}
%
\institute{University of Science and Technology of China, Hefei, China\\ \email{\{yuntaowei,zyz2020,zhangsy2023\}@mail.ustc.edu.cn;\\zhangh@ustc.edu.cn}}

\maketitle

\begin{abstract}
Multimodal depression detection is an important research topic that aims to predict human mental states using multimodal data. 
Previous methods treat different modalities equally and fuse each modality by na\"ive mathematical operations without measuring the relative importance between them, which cannot obtain well-performed multimodal representations for downstream depression tasks. In order to tackle the aforementioned concern, we present a \textbf{C}ross-modal \textbf{A}ttention \textbf{N}etwork with \textbf{A}daptive \textbf{M}ulti-modal \textbf{R}ecurrent \textbf{F}usion (CANAMRF) for multimodal depression detection. CANAMRF is constructed by a multimodal feature extractor, an Adaptive Multimodal Recurrent Fusion module, and a Hybrid Attention Module. Through experimentation on two benchmark datasets, CANAMRF demonstrates state-of-the-art performance, underscoring the effectiveness of our proposed approach.

\end{abstract}

\section{Introduction}

Depression stands as a prevalent psychiatric disorder while preserving implicit symptoms. Patients haunted by depression often resist timely treatment for fear of misunderstanding from other people, which casts tremendous shade on both their physical and mental health. Recently, a significant amount of research attention has been directed towards the development of multimodal depression assessment systems. These systems leverage diverse cues from text, audio, and video to evaluate depression levels and facilitate diagnostic processes. 

However, previous works either focus on single-modality text information or treat each modality equally, and then propose various fusion methods on this basis.  Gong et al. \cite{topic_modeling_method} utilized topic modeling to partition interviews into segments related to specific topics. Additionally, a feature selection algorithm was employed to retain the most distinctive features. Hanai et al. \cite{Mlut-model_LSTM} analyzed data from 142 individuals undergoing depression screening. They employed a Long-Short Term Memory neural network model to detect depression by modeling interactions with audio and text features. Yuan et al. \cite{MMFF} proposed a multimodal multiorder factor fusion method to exploit high-order interactions between different modalities. This fusion method, which does not discriminate between each modality, cannot well mine the main features that are more effective for depression detection. At the same time, the traditional audio, text, and vision features have not been better in making the category distinction.

In response to these limitations, we introduce a \textbf{C}ross-modal \textbf{A}ttention \textbf{N}etwork with \textbf{A}daptive \textbf{M}ulti-modal \textbf{R}ecurrent \textbf{F}usion. CANAMRF first extracts  features of four modalities, including textual, acoustic, visual, and the newly proposed sentiment structural modalities, by specific feature extractors separately, then fuses textual features with the other three features through AMRF module. Finally, it utilizes a hybrid attention module to generate distinguishable multimodal representations for subsequent depression detection tasks.

Our primary contributions can be succinctly summarized as follows: 
\begin{itemize}
    \item [1)]
    We introduce sentiment structural modality as a supplementary modality as a means to augment the performance of multimodal depression detection.
    \item [2)]
    We present an innovative approach to modality fusion called Adaptive Multimodal Recurrence Fusion (AMRF). It can dynamically adjust the fusion weights of different modalities, which realizes the trade-off between modalities and has excellent performance.
    \item [3)]
    We build a hybrid attention module, which combines cross-modal attention and self-attention mechanisms, to generate representative multimodal features. Extensive experiments and comprehensive analyses are provided to showcase the efficacy of our suggested method.
\end{itemize}

\section{Methodology}
\begin{figure}[h]
\includegraphics[width=0.9\textwidth]{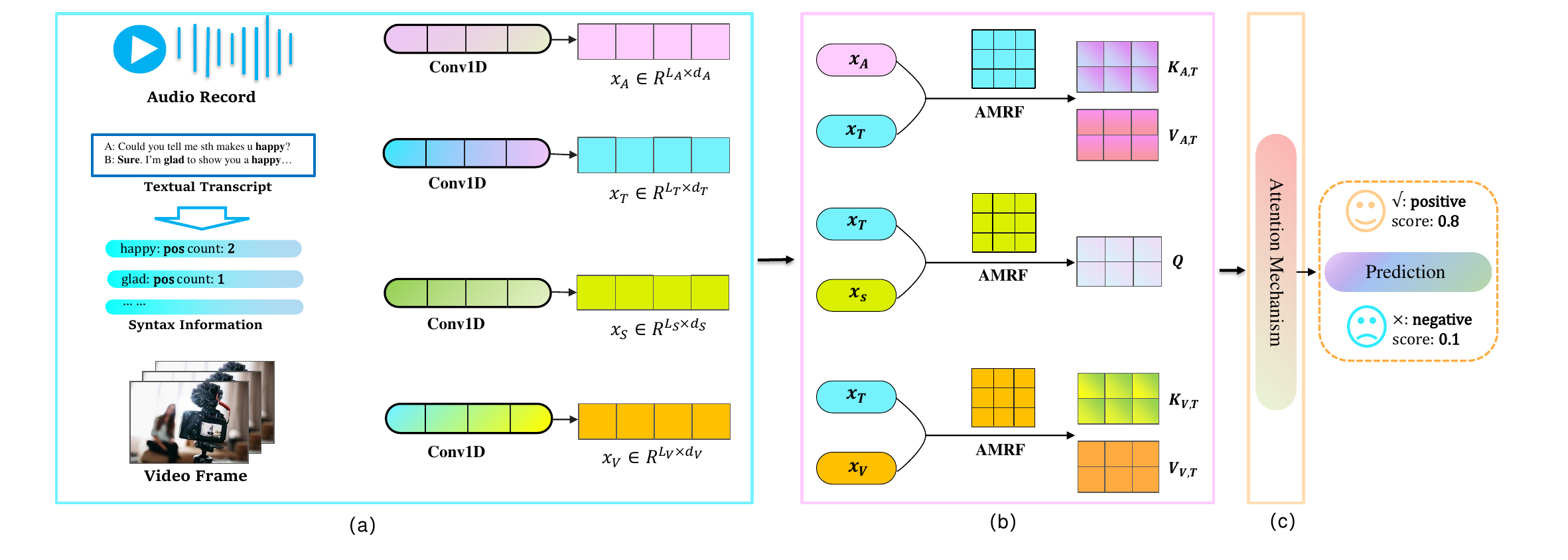}
\caption{The overall framework of CANAMRF. (a) Feature extraction procedure for multiple modalities; (b) Fusion of modalities through AMRF module; (c) Hybrid Attention Mechanism.}
\label{figure1}
\end{figure}

In this section, we elucidate the specifics of CANAMRF, illustrated in Figure \ref{figure1}. 

\subsection{Feature Extractor}
\label{sub31}
We use specific open-source toolkits and pretrained models, including OpenFace \cite{OpenFace}, OpenSMILE \cite{OpenSMILE}, and BERT \cite{BERT}, for extracting features for textual, visual, and acoustic modalities. In addition, we also introduce a novel high-level semantic structural modality, which consists of five word-level features and three sentence-level features. All features are passed into a 1D temporal convolutional layer to be reshaped into vectors of the same dimention $d$ for subsequent depression detection tasks.

\subsection{Adaptive Multi-modal Recurrent Fusion}
\begin{figure}[htbp]
    \centering
    \begin{subfigure}[t]{0.48\linewidth}
        \centering
        \includegraphics[width=\linewidth]{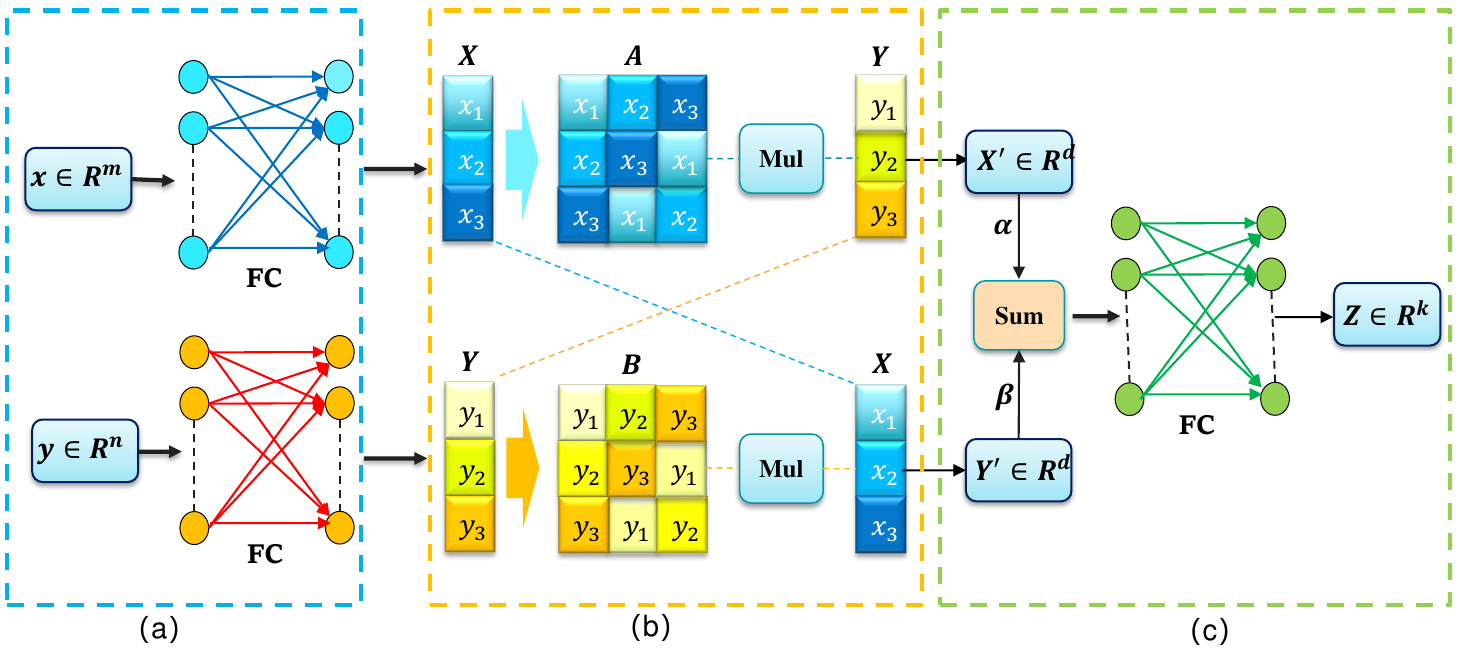}
        \caption{}
    \end{subfigure}
    \hfill
    \begin{subfigure}[b]{0.48\linewidth}
        \centering
        \includegraphics[width=\linewidth]{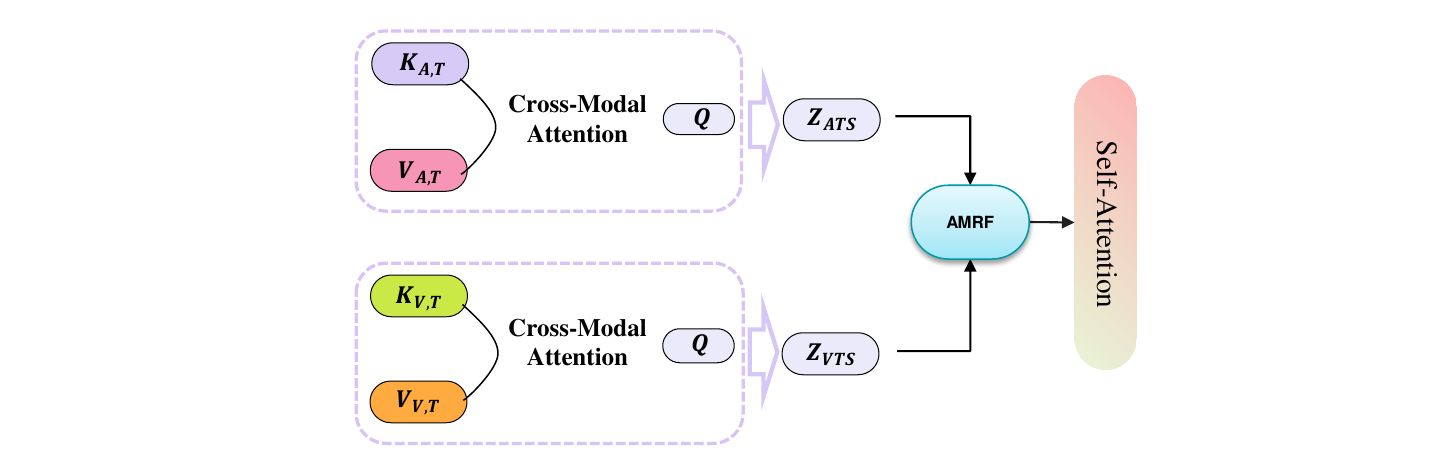}
        \caption{}
    \end{subfigure}
    \caption{Subfigure (a): the framework of AMRF module. (a) Features from different modalities are projected into a same low-dimensional space by fully-connected layers; (b) The low-dimensional features are further processed by \textit{Recur} operation; (c) Features are fused according to the adaptive fusion mechanism, and transformed via fully-connected layers to obtain the the final representation; Subfigure (b): the framework of Hybrid Attention Module.}
    \label{figure2}
\end{figure}

The detailed framework of Adaptive Multi-modal Recurrent Fusion (AMRF) module is illustrated in Fig. \ref{figure2}(a).

Following Wu et al. \cite{Texture_centered}, we always fuse textual features with features of other modalities (probably visual, acoustic, and sentiment structure), because of the predominance of textual features. Given two feature vectors of different modalities, for example acoustic characteristics $x \in \mathbb{R}^m$ and textual characteristics $y \in \mathbb{R}^n$, we first map them in a common dimension space $d$ utilizing two projection matrices $W_1 \in \mathbb{R}^{d \times m}$ and $W_2 \in \mathbb{R}^{d \times n}(d \leq \min(m,n))$ by $X = xW_1^\mathsf{T}, \quad Y = yW_2^\mathsf{T}$,
where $W_1^\mathsf{T}$ and $W_2^\mathsf{T}$ are the transpose of $W_1$ and $W_2$, respectively.

Then we construct recurrent matrices $A \in \mathbb{R}^{d \times d}$ and $B \in \mathbb{R}^{d \times d}$ using projected vectors $X \in \mathbb{R}^{d}$ and $Y \in \mathbb{R}^{d}$ by $A = Recur(X), \quad B = Recur(Y)$, where $Recur(\cdot)$ is the operation to construct the recurrent matrix from a vector as visualized in part (b) of Fig.\ref{figure2}(a), which illustrates that all rows of $A$ forms a circular permutation of vector $X$. In order to fully fuse the elements in the projected vector and recurrent matrix, each row vector of the recurrent matrix is multiplied with the projected vector and then added to the average, as shown in Eq.\eqref{equ3}:
\begin{equation}\label{equ3}
X^{'} = \frac{1}{d} \sum_{i=1}^{d} a_{i} \odot A, \quad Y^{'} = \frac{1}{d} \sum_{i=1}^{d} b_{i} \odot B ,
\end{equation}
where $a_i \in \mathbb{R}^d$ and $b_i \in \mathbb{R}^d$ are the $i$th row vectors of $A$ and $B$, respectively. $\odot$ denotes the elementwise multiplication.

The final fused features are obtained by $Z = (\alpha  X^{'} + \beta  Y^{'}) W_3^\mathsf{T}$, where $\alpha$ and $\beta$ ($0 \leq \alpha, \beta \leq 1$) are two learnable weight parameters, and $W_3 \in \mathbb{R}^{d \times k}$ is a projection matrix.

\subsection{Hybrid Attention Module}
In this subsection, we illustrate the hybrid attention module, whose framework is shown in Fig. \ref{figure2}(b). The hybrid attention module consists of cross-modal attention module, AMRF, and self-attention module.

Following Wu et al. \cite{Texture_centered}, we conduct the attention operation between textual modality and the remaining three modalities. Let $X_{\alpha\beta}$ be the fusion result from modality $\alpha$ and modality $\beta$. With the AMRF module, the fusion process can be formulated as: 
\begin{equation}\label{equ6}
    Q = AMRF(X_S, X_T) ,
\end{equation}
\begin{equation}\label{equ7}
    K_{VT} = V_{VT} = X_{VT} = AMRF(X_V, X_T) ,
\end{equation}
\begin{equation}\label{equ8}
    K_{AT} = V_{AT} = X_{AT} = AMRF(X_A, X_T) ,
\end{equation}
The cross-modal attention mechanism can be formulated as:

\begin{equation}\label{equ9}
    Z_{ATS} = CMA(Q,K_{AT},V_{AT}) = softmax\left (\frac{QK_{AT}^\mathsf{T}}{\sqrt{d_k} } \right ) V_{AT} ,
\end{equation}
\begin{equation}\label{equ10}
    Z_{VTS} = CMA(Q,K_{VT},V_{VT}) = softmax\left (\frac{QK_{VT}^\mathsf{T}}{\sqrt{d_k} } \right ) V_{VT} ,
\end{equation}
where $Z_{ATS}$ and $Z_{VTS}$ are the output of Cross-Modal Attention. In order to fully integrate four features and have a certain adaptive weight ratio, we pass $Z_{ATS}$ and $Z_{VTS}$ through the AMRF module, followed by a Self-Attention module to get the final fused feature, as shown in Eq. \eqref{equ11}:
\begin{equation}\label{equ11}
    Z_f = Self-Attention(AMRF(Z_{ATS}, Z_{VTS})) .
\end{equation}

\subsection{Training Objective}
The fused multimodal feature $Z_f$ is flattened and then fed into the fully-connected layers to predict whether a subject has depression or not: $\hat{y} = \sigma(FC(Flatten(Z_f)))$,
where $Flatten(\cdot)$ is the flatten operator, $FC(\cdot)$ is the fully-connected layers, and $\sigma(\cdot)$ is an activation function.

We use the Focal loss to train the model: $\mathcal{L}_{fl}=-(1-\tilde{y})^{\gamma} \log(\tilde{y})$,
where $\tilde{y}$ is the estimated probability of being a positive class, and $\gamma \geq 0$ is a tuning parameter.

\section{Experiments}
In this section, we assess  the performance of the proposed CANAMRF using two benchmark datasets, including the Chinese Multimodal Depression Corpus (CMDC)\cite{CMDC} and EATD-Corpus\cite{GRU_BiLSTM} which are frequently used in previous work.

\subsection{Baselines}
For CMDC and EATD-Corpus, we compare CANAMRF with the following machine learning models: (1) Linear kernel support vector machine (SVM-Linear); (2) SVM based on sequential minimal optimization \cite{SMO} (SVM-SMO); (3) Logistic regression; (4) Na\"ive Bayes; (5) Random Forest; (6) Decision Tree; and the following deep learning models: (1) Multimodal LSTM \cite{Mlut-model_LSTM}; (2) GRU/BiLSTM-based model \cite{LSTM-CRF}; (3) Multimodal Transformer \cite{MulT}; (4) TAMFN \cite{TAMFN}.

\begin{table}[!ht]
    \centering
    \caption{Performance comparison between baseline models and CANAMRF.}
    \resizebox{\linewidth}{!}{
    \begin{tabular}{ccccccccccccccccc}
    \toprule
    & \multirow{2}{*}{\textbf{Model}}  & \multicolumn{3}{c}{ \textbf{CMDC(AT)}} & \multicolumn{3}{c}{ \textbf{CMDC(ATV)}}  & \multicolumn{3}{c}{ \textbf{EATD(T)}} & \multicolumn{3}{c}{ \textbf{EATD(A)}}  & \multicolumn{3}{c}{ \textbf{EATD(AT)}} \\
    \cmidrule(lr){3-5} \cmidrule(lr){6-8} \cmidrule(lr){9-11}  \cmidrule(lr){12-14} \cmidrule(lr){15-17}
     & & $P.$ & $R.$ & $F_1$& $P.$ & $R.$ & $F_1$& $P.$ & $R.$ & $F_1$ & $P.$ & $R.$ & $F_1$& $P.$ & $R.$ & $F_1$ \\
    \midrule
    &  SVM-Linear & 0.90 & 0.91 & 0.91 & 0.91 & 0.91 & 0.91 & 0.48 & \textbf{1.00} & 0.64 & 0.54 & 0.41 & 0.46 & - & - & -  \\
    & SVM-SMO & 0.92 & 0.91 & 0.91 & 0.91 & 0.89 & 0.89 & - & - & - & - & - & - & - & - & -\\
    & Na\"ive Bayes & 0.91 & 0.89 & 0.89 & 0.84 & 0.84 & 0.84 & - & - & - & - & - & - & - & - & -\\ 
    &Random Forest & - & - & - & - & - & - & 0.61 & 0.53 & 0.57 & 0.48 & 0.53 & 0.50 & - & - & -\\
    &Logistic Regression &  0.92 & 0.91 & 0.91 & 0.82 & 0.82 & 0.82 & - & - & -  & - & - & - & - & - & -\\
    &Decision Tree & - & - & - & - & - & - & 0.59 & 0.43 & 0.49 & 0.47 & 0.44 & 0.45 & - & - & -\\
    &Multi-modal LSTM & - & - & - & - & - & - & 0.53 & 0.63 & 0.57 & 0.44 & 0.56 & 0.49 & 0.49 & 0.67 & 0.57\\
    &GRU/BiLSTM-based Model & \textbf{0.97} & 0.91 & 0.94 & 0.87 & 0.89 & 0.88 & \textbf{0.65} & 0.66 & \textbf{0.65} & \textbf{0.57} & \textbf{0.78} & \textbf{0.66} & 0.62 & 0.84 & 0.71\\
    &MulT & 0.87 & 0.96 & 0.91 & \textbf{0.97} & 0.85 & 0.91 & - & - & - & - & - & - & - & - & -\\
    &TAMFN & - & - & - & - & - & - & - & - & - & - & - & - & 0.69 & \textbf{0.85} & 0.75\\
    \midrule
    &\textbf{CANAMRF} & 0.94 & \textbf{0.97} & \textbf{0.95} & 0.95 & \textbf{0.93} & \textbf{0.93} & - & - & - & - & - & - & \textbf{0.71} & 0.83 & \textbf{0.77}\\
    \bottomrule
    \end{tabular}
    }
    \label{main_results}
\end{table}

\subsection{Main Results}
Table \ref{main_results} displays the performance comparison of the benchmark models and the CANAMRF model on the CMDC and EATD datasets. For both CMDC and EATD datasets, CANAMRF consistently outperforms the state-of-the-art baselines on $F_1$ scores, which highlights the effectiveness of CANAMRF, both in unimodal and multimodal depression detection tasks.

\section{Conclusion}
In this article, we present CANAMRF, a comprehensive framework consisting of three key components. First, we introduce an effective sentiment structural modality as a supplementary modality to enhance the performance of multimodal depression detection tasks. Next, we treat the textual modality as the dominant modality and fuse it with the remaining three modalities using the AMRF module. Finally, we process the fused features using a hybrid attention module to obtain distinct multimodal representations. The experimental results demonstrate the high effectiveness and promising potential of CANAMRF in the detection of depression.
\\

\noindent\textbf{Acknowledgment}. This research was partially supported by National Natural Science Foundation of China (No. 12171451) and Anhui Center for Applied Mathematics.
\bibliographystyle{splncs04}
\bibliography{refs}

\end{document}